\documentclass[letterpaper, 10 pt, conference]{ieeeconf}  
\usepackage{cite}
\usepackage{graphicx}
\usepackage{dblfloatfix}

\usepackage{leftidx}

\usepackage{bm} 
\usepackage{booktabs} 

\usepackage{array}

\let\labelindent\relax
\usepackage{enumitem}

\usepackage{algorithmic}
\usepackage{algorithm}
\usepackage{amssymb}
\usepackage{mathtools}
\usepackage{tabularx}
\usepackage{subfigure}
\usepackage{caption}

\usepackage[colorlinks,
            linkcolor=black,
            anchorcolor=black,
            citecolor=black,
            ]{hyperref}

\begin{document}            
\newcommand{\etal}{\textit{et al. }}
\renewcommand*{\figureautorefname}{Fig.}

\IEEEoverridecommandlockouts                          

\title{\LARGE \bf
Teach Biped Robots to Walk via Gait Principles and Reinforcement Learning with Adversarial Critics}

\author{Kuangen~Zhang$^{1, 2}$, Zhimin Hou$^{3}$, Clarence W.~de Silva$^{2}$, Haoyong Yu$^{3}$, and~Chenglong~Fu$^{1}$
\thanks{*This work was supported by the National Key R\&D Program of China under Grant 2018YFB1305400, National Natural Science Foundation of China under Grant U1613206 and Grant 61533004, Guangdong Innovative and Entrepreneurial Research Team Program under Grant 2016ZT06G587, Shenzhen and Hong Kong Innovation Circle Project (Grant No. SGLH20180619172011638), and Centers for Mechanical Engineering Research and Education at MIT and SUSTech.}
\thanks{$^{1}$K. Zhang and C. Fu are with the Department of Mechanical and Energy Engineering, Southern University of Science and Technology, Shenzhen 518055, China (Corresponding author: Chenglong Fu: fucl@sustech.edu.cn)}%
\thanks{$^{2}$K. Zhang and C. W. de Silva are with the Department of Mechanical Engineering, The University of British Columbia, Vancouver V6T1Z4, Canada.}%
\thanks{$^{3}$Z. Hou and H. Yu are with the Department of Biomedical Engineering, National University of Singapore, Singapore 117583.}%
\thanks{Code and data: \url{https://github.com/KuangenZhang/ATD3}}
}

\markboth{IEEE Robotics and Automation Letters}%
{Shell \MakeLowercase{\textit{et al.}}: Bare Demo of IEEEtran.cls for IEEE Journals}

\maketitle

\begin{abstract}
Controlling a biped robot to walk stably is a challenging task considering its nonlinearity and hybrid dynamics. Reinforcement learning can address these issues by directly mapping the observed states to optimal actions that maximize the cumulative reward. However, the local minima caused by unsuitable rewards and the overestimation of the cumulative reward impede the maximization of the cumulative reward. To increase the cumulative reward, this paper designs a gait reward based on walking principles, which compensates the local minima for unnatural motions. Besides, an Adversarial Twin Delayed Deep Deterministic (ATD3) policy gradient algorithm with a recurrent neural network (RNN) is proposed to further boost the cumulative reward by mitigating the overestimation of the cumulative reward. Experimental results in the Roboschool Walker2d and Webots Atlas simulators indicate that the test rewards increase by 23.50\% and 9.63\% after adding the gait reward. The test rewards further increase by 15.96\% and 12.68\% after using the ATD3\_RNN, and the reason may be that the ATD3\_RNN decreases the error of estimating cumulative reward from 19.86\% to 3.35\%. Besides, the cosine kinetic similarity between the human and the biped robot trained by the gait reward and ATD3\_RNN increases by over 69.23\%. Consequently, the designed gait reward and ATD3\_RNN boost the cumulative reward and teach biped robots to walk better.
\end{abstract}


\section{Introduction}
Biped robots are able to overcome obstacles in complex environments because they choose the contact point on the terrain freely \cite{hwangbo_learning_2019}. 
However, it is still a challenge to control the biped robot to walk stably. Unlike the robot manipulator, which is fully-actuated and limited in a specific workspace, the biped robot is an underactuated, high dimensional, non-linear, and hybrid system, and it needs to interact with complex environments in real-time \cite{castillo_reinforcement_2019}.

To realize the locomotion control of the biped robots, most existing research utilized model-based methods, which include a feedforward trajectory planner and a feedback sensory controller \cite{fu_gait_2008, de_silva_sensor_2017}. The trajectory planner optimizes the future trajectory of the swing foot and the torso under the constraints of environments, kinematics, and stability. 
Considering the high degrees of freedom of the biped robots, a linear inverted pendulum model \cite{martin_experimental_2017} can be combined with the zero-moment point (ZMP) criteria to optimize the gait trajectory \cite{vukobratovic_zero-moment_2004}. However, the designed trajectory based on the simplified model may be suboptimal and unnatural. To generate the whole-body motion more accurately, the full-body dynamics of the robot can be estimated in a simulation environment (e.g. MuJoCo) and the model predictive control (MPC) method can be utilized to optimize the future trajectory in real-time \cite{erez_integrated_2013}. 
The optimized trajectories are then converted to the joint trajectories using the inverse kinematics of the robot.
Although the model-based methods have achieved an acceptable performance for controlling biped robots, there are still several limitations. The model-based methods are based on the accurate dynamic model, which is difficult to obtain for real robots. Additionally, the online trajectory optimization is a constrained nonlinear optimization problem and requires a high computational cost, which limits the bandwidth of the controller.

Reinforcement learning (RL), a type of data-driven method, is able to address the aforementioned problems \cite{castillo_reinforcement_2019}. On the one hand, the RL trains the robot to directly learn controllers from experiences without the requirement of the robotic model \cite{hou_knowledge-driven_2018}. On the other hand, a neural network can directly map the states of environments and robots to the actions \cite{xu_feedback_2019, zhang_environmental_2019, li_using_2019, lobos-tsunekawa_visual_2018}, which is more efficient than optimizing the trajectory in real-time. The core of the RL is to train the robot to take the action that maximizes the expected cumulative reward. The reward represents the objective and inappropriate rewards may lead to the local minima. Besides, the cumulative reward includes the reward in the future, which cannot be measured directly. To maximize the cumulative reward, the situations of local minima should be reduced and the error of estimating the cumulative reward should be decreased.

Designing an appropriate reward needs to analyze the specific task and is usually a case study, and thus many end-to-end RL methods do not consider the specific characteristics of each control problem and directly utilize the default reward provided by the simulators \cite{lillicrap_continuous_2015, haarnoja_soft_2018, fujimoto_addressing_2018}. However, these methods may fall into the local minima and generate an unnatural motion that is unacceptable for robots, especially wearable robots \cite{wen_online_2019}. An intuitive approach to reduce the situations of local minima is to incorporate some prior knowledge \cite{peng_deepmimic:_2018}. Peng \etal adopted the RL to train the simulated robots to imitate a wide variety of captured motion clips, including backflip and cartwheel, and accomplish user-specified goals \cite{peng_deepmimic:_2018}. Xie \etal utilized the deep RL to control the Cassie robot to imitate a designed reference motion \cite{xie_iterative_2019}. However, the imitation learning needs to input a reference motion to the neural network to estimate the next actions, which limits the motion of the robot and just solves the trajectory tracking problem. As discussed before, planning a trajectory is also problematic and it is still difficult to control the robot without the reference trajectory. Therefore, the reward based on the online reference motion is not appropriate. It remains a problem to design a suitable reward for the biped robot.

As for the problem of estimating the cumulative reward, the actor-critic based RL is designed recently to address it \cite{lillicrap_continuous_2015,haarnoja_soft_2018, fujimoto_addressing_2018}. The critic network estimates the Q-value, which is the expected cumulative reward of the input states and actions, while the actor-network generates the action that maximizes the Q-value. Under such a greedy strategy, the critic may overestimate the Q-value and result in a suboptimal actor \cite{fujimoto_addressing_2018}. To address this issue, Fujimoto \etal designed a Twin Delayed Deep Deterministic policy gradient algorithm (TD3) with two critics to estimate the Q value more accurately \cite{fujimoto_addressing_2018}. However, the actor of the TD3 only maximizes the Q value of one critic, which may affect the accuracy of estimating the Q value. 


Given the possible limitations of the general end-to-end RL and imitation learning on controlling the biped robot, a natural question is how to design an appropriate reward for the biped robot without a reference motion. The answer in this paper is to design a gait reward based on walking principles, including the gait cycle and the stance phase, to evaluate the historical gaits of the robot. To further increase the cumulative reward, this paper also designs an Adversarial TD3 with a recurrent neural network (ATD3\_RNN) to consider the relationship between two critics and maximize the mean of Q-values estimated by two critics. The ATD3\_RNN is first evaluated using a biped robot (RoboschoolWalker2d-v1) with six degrees of freedom based on the Roboschool, which is an open-source robot simulator developed by OpenAI. This paper also builds an Atlas robot with reinforcement learning interfaces in the Webots, which is a professional robotics simulator, to further evaluate the proposed method.

The key contributions of the present paper include:
\begin{enumerate}
    \item Propose a gait reward based on walking principles, which assist the robot to avoid some situations of the local minima.
    \item Design a pair of adversarial critics and maximize the mean of their Q-values using a recurrent neural network to further improve the learned policy.
\end{enumerate}

The rest of this paper is organized as follows. 
The biped walking problem and designed reward system are introduced in section \ref{sec:Problem}. Section \ref{sec:AdversarialTD3} proposes our ATD\_RNN algorithm. Section \ref{sec:Results} gives the experimental setup, results, and corresponding discussions. Section \ref{sec:Conclusion} concludes the paper.

\section{Problem Formulation}
\label{sec:Problem}
In this section, the biped robot model and its observable states and controllable actions are described. Then the objective and the default reward of controlling such a biped robot are introduced. Finally, this paper designs a gait reward, including the reward of the double support period, the gait cycle, and the crossover gait.

\subsection{Biped robot model}
\begin{figure*}[h!]
    \centering
    \vspace*{0.5em}
    \includegraphics[width=\textwidth]{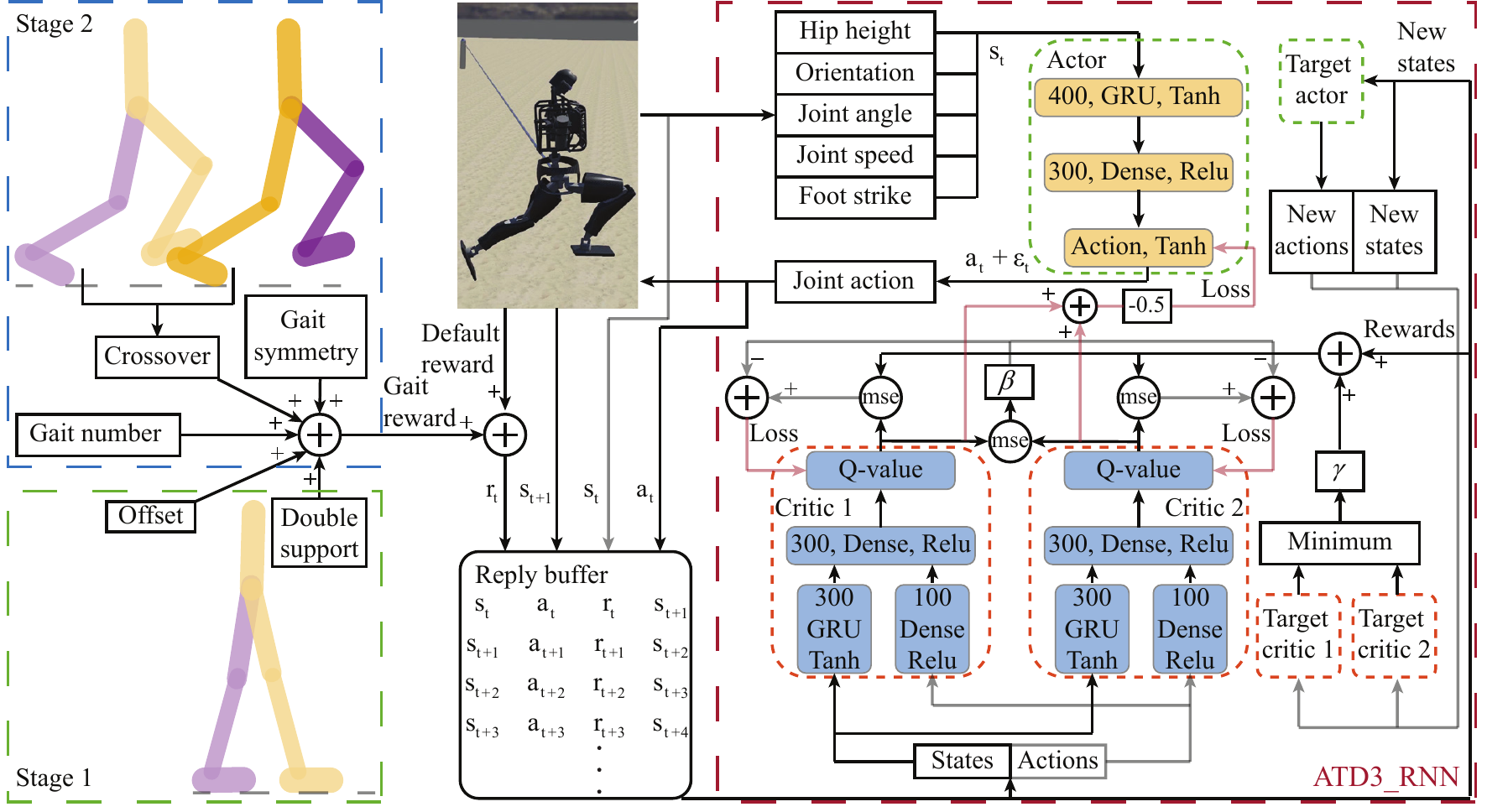}
    \caption{Overview of the ATD3\_RNN and the gait reward. The ATD3\_RNN has an actor and two adversarial critics consist of multiple dense layers and a gated recurrent unit (GRU), which is a kind of RNN. The adversarial critics estimate Q-values and maximize the mean squared error (mse) between each other. The input of the GRU is a state sequence of a fixed length $\bm{s}_t$ that includes state vectors in a time sequence and the end vector of $\bm{s}_t$ is the current state vector. The actor receives $\bm{s}_t$ and outputs an optimal action $\bm{a}_t$ to maximize the mean of Q-values. The exploration noise $\bm{\varepsilon}_t$ is added on the $\bm{a}_t$ to avoid the local minima. The ATD3\_RNN degrades to ATD3 if the GRU is replaced by a dense layer and the sequence length is 1. The simulator generates a default reward and updates a new state sequence $\bm{s}_{t+1}$ based on the $\bm{s}_t$ and $\bm{a}_t$. The default reward is added with a designed gait reward to the reward $r_t$. The gait reward includes rewards which encourage the robot to walk (stage 1) and teach the robot to learn walking principles (stage 2).  The $\bm{s}_t$, $\bm{a}_t$, $r_t$, and $\bm{s}_{t+1}$ are combined in a tuple and saved in a buffer, which provides the samples for training the actor and critics. $\beta$ and $\gamma$ are two discount factors.} 
    \label{fig:1_overview} 
\end{figure*}

This paper focuses on the planar walking, and thus the Roboschool Walker2d used in this paper is a 2D biped robot. The Atlas robot built in the Webots is also connected with a boom, which enforces the Atlas robot to walk in the sagittal plane (see \autoref{fig:1_overview}). The 2D biped robot usually has seven links: a torso, a right/left thigh, a right/left shank, and a right/left foot. Six actuated joints, including a right/left hip, a right/left knee, and a right/left ankle, connect the above links. The joints can only rotate in the sagittal plane. There are 22 observable states in the state vector $\bm{s}_t$ for both the Walker2d and the Atlas robot:
\begin{equation}
\begin{split}
   \bm{s}_t = [& z_t - z_0, \cos{(\hat{q}_t^y - q_t^y)}, \sin{(\hat{q}_t^y - q_t^y)}, v_t^x, v_t^y,  \\
   &v_t^z, q_t^r, q_t^p, q_t^{rh}, \dot{q}_t^{rh}, \dots, q_t^{la}, \dot{q}_t^{la}, f_t^r, f_t^l],
\end{split}
\end{equation}
where $z_t - z_0$ represents the deviation between the current hip height $z_t$ and initial hip height $z_0$. $q_t^r$ (roll), $q_t^p$ (pitch), $q_t^y$ (yaw), and $\hat{q}_t^y$ (desired yaw) are the Euler angles of the torso and the target yaw. For the 2D biped robot, only $q_t^p$ is not zero. The velocities of the torso along the local $x$, $y$, and $z$ direction are represented by $v_t^x$, $v_t^y$, and $v_t^z$. The joint angles and angular velocities of right hip ($rh$), right knee ($rk$), right ankle  ($ra$), left hip  ($lh$), left knee ($lk$), and left ankle ($la$) are denoted by $q_t^j$ and $\dot{q}_t^j$, respectively. $f_{t}^r$ (right) and $f_{t}^l$ (left) represent whether the foot contacts the ground.

The controllable actions of the 2D biped robot are torques $\tau_t^j$ of six joint motors for the Roboschool Walker2d and joint angles $q_t^j$ of six joint motors for the Webots Atlas. The ranges of hip, keen, and ankle are changed to $[-25^\circ, 115^\circ]$, $[0^\circ, 150^\circ]$, and $[-45^\circ, 45^\circ]$, respectively. The positive angle represents flexion or dorsiflexion.

\subsection{Reinforcement learning}
\label{subsec:RL}
The basic idea of the RL algorithm is to train the robot to learn a behavior by maximizing the expected reward. 
At each time step $t$, the robot selects the ``optimal" action $\bm{a}_t \in \mathcal{A}$ according to the current state $\bm{s}_t \in \mathcal{S}$ and the policy $\pi(\bm{a}_t|\bm{s}_t): \mathcal{S}\rightarrow\mathcal{A}$. 
The robot will receive a reward $r_t$ from environment to evaluate the state-action pair and transform to the next state $\bm{s}_{t+1}$. 
Therefore, the objective is to maximize the expected cumulative reward $J$ as: 

\begin{equation}\label{eq:expected_return}
    J = \mathbb{E}_{\bm{s}_{t+1} \thicksim p_{\pi}(\bm{s}_t), \bm{a}_t \thicksim \pi} \left[\sum_{t=0}^{T-1} \gamma^t r_{t}(\bm{s}_t, \bm{a}_t) \right],
\end{equation}
where $\gamma$ is a discount factor of the short-term reward.

\subsubsection{Default rewards}
The default objective of the biped robot is to walk as fast as possible and remain stable in a constant time period. Roboschool simulator provides a default reward, which is set as $r_t^d$ for a time step $t$:
\begin{equation}\label{eq:default_reward}
\begin{split}
   r_t^d &= r_t^a + r_t^p + r_t^\tau + r_t^l + r_t^c, 
\end{split}
\end{equation}
where $r_t^a$ is a positive constant if the pose of the torso is natural otherwise is a negative constant; $r_t^p$ is proportional to the forward speed; $r_t^\tau$ is inverse proportional to the power and torque of the joint motors; $r_t^l$ and $r_t^c$ punish the agent when the joint angle approaches the limit or the self collision happens.

\subsubsection{Gait reward}
Experimental results show that the robot trained by the default reward sometimes inclines to stand rather than walk. The reason may be that standing is a stable status and robot can get at least an alive reward. Conversely, walking is unstable in most situations unless the robot has learned a suitable walking pattern that achieves a dynamic balance. To avoid the local minima of the standing behavior and other unnatural motions, an appropriate reward needs to be designed. Previous researchers have trained the robot to track a desired reference motion \cite{peng_deepmimic:_2018}, but this strategy limits the motion of the robot and needs to input a planned trajectory. In this paper, only some principles of human gait are utilized to evaluate the historical gait of the robot, and thus the robot can move freely without tracking a trajectory. The gait reward is based on a complete gait and is a sparse reward, which adds the same reward in the evaluated gait. A complete gait starts at the right heel strike and ends at the next right heel strike and the gait cycle should be higher than 25 time steps. The gait reward $r^g$ is defined as: 
\begin{equation}
\begin{split}
   r^g = &r^s + r^n + r^{lhs} + r^{cg} + r^{gs}, 
\end{split}
\end{equation}
where different sub gait rewards describes different principles of human gait, which are described in the \autoref{tab:gait_reward} and in the following paragraphs.

\begin{table}[htbp]
\vspace*{0.5em}
\caption {\label{tab:gait_reward} Definitions of sub gait rewards.}
\renewcommand{\arraystretch}{1.5} 
\begin{center}
\resizebox{0.99\columnwidth}{!}{
\begin{tabular}{l l}
\toprule
Principles & Reward equation\\
\midrule
Double support period &$r^s = 
\begin{cases} 
-2.0, &\text{if } t^\text{s} > 100,\\
0, &\text{otherwise}.
\end{cases}
$\\
Gait number &$r^n = 0.05 * (n_{t+500} - n_t)$\\
Left heel strike &$r^{lhs} = 0.2 * \Big(1-\text{tanh}\big(({t^{lhs}}/{T^g}-0.5)^2\big)\Big)$\\
Crossover gait &
$\begin{aligned} 
r^{cg} = 0.05 * \big(
&\text{tanh}(q_{t^{rhs}}^{rh} - q_{t^{rhs}}^{rk}) ~+\\
&\text{tanh}(-q_{t^{lhs}}^{rh} + q_{t^{lhs}}^{rk}) ~+\\
&\text{tanh}(-q_{t^{rhs}}^{lh} + q_{t^{rhs}}^{lk}) ~+\\
&\text{tanh}(q_{t^{lhs}}^{lh} - q_{t^{lhs}}^{lk})
\big)
\end{aligned}
$\\
Gait symmertry &$r^{gs} = (0.2/3)*\sum_{j = 1}^3{({\bm{q}_j^r}^T \cdot \bm{q}_j^l)/(\|\bm{q}_j^r\|\|\bm{q}_j^l\|)}$\\
\end{tabular}
}
\end{center}
\end{table}
The first objective is to train the robot to walk stably rather than stand, and thus rewards in the stage 1 include:
\begin{itemize}[leftmargin=1em]
    \item Offset reward (-0.5): It is used to decrease the alive reward at each time step to avoid remaining standing.
    \item The reward of the double support period $r^s$: The double support period $t^\text{s}$ is defined as the continuous time steps when two feet are both in contact with the ground. A double support period only lasts for about 10\% of a complete gait \cite{whittle_gait_2014}, and thus $r^s$ is negative and the episode is finished if the double support period $t^\text{s}$ is higher than a threshold.
\end{itemize}

\textcolor{black}{
After the robot finishes two complete gaits, the robot starts to learn walking principles. The rewards in the stage 2 inclue: 
\begin{itemize}[leftmargin=1em]
    \item The reward of gait number $r^n$: It is proportional to the number of future gaits within 500 timesteps. $r^n$ is added because walking a long distance requires the robot to walk stably for multiple gaits. Walking can be seen as a sequential model \cite{zhang_sequential_2019}, and the end state of the last gait is the initial state of the current gait. A suitable initial state will promote the robot to achieve the gait stably. 
    \item The reward of the left heel strike $r^{lhs}$: It is to encourage the time of the left heel strike $t^{lhs}$ to approach the middle time of a gait cycle $T^g$ to imporve the gait symmetry \cite{whittle_gait_2014}. The Tanh function is used to map the reward from (-infinity, +infinity) to (-1, 1).
    \item The reward of the crossover gait $r^{cg}$: The leading leg of a human changes in a gait cycle \cite{whittle_gait_2014}, and $r^{cg}$ is related to the relative position between the ankle and the hip, which is calculated using the joint angles of the hip and knee, when the heel strike happens.
    \item The reward of the gait symmetry $r^{gs}$: There is a gait symmetry betwen two sides for humans \cite{whittle_gait_2014}, and thus $r^{gs}$ is proportional to the averaged cosine similarity between the joint angular curves of the right leg $q_j^r$ and those of the left leg $q_j^l$. The joint angular curves start at the ipsilateral heel strike and end at the next ipsilateral heel strike.
\end{itemize}
}

While training the robot, the overall reward is set as $-0.5 + r^d_t + r^g$. To objectively compare with the state-of-art result, the reward remains the default reward $r^d_t$ while testing the robot.

\section{ATD3\_RNN Algorithm}
\label{sec:AdversarialTD3}
The framework of the proposed Adversarial Twin Delayed Deep Deterministic policy gradient algorithm with a recurrent neural network (ATD3\_RNN) is demonstrated in \autoref{fig:1_overview}. ATD3 is based on the TD3 algorithm \cite{fujimoto_addressing_2018, silver_deterministic_2014} and adds an adversarial loss between two critics and calculate the mean of their Q-values to further mitigate the problem of overestimating Q-value. As explained in \ref{subsec:RL}, the actor is a policy network $\pi_{\phi}(\bm{a}_t|\bm{s}_t)$ with parameters $\phi$, which generates an optimal action $\bm{a}_t$ to maximize the expected reward. Two adversarial critics $Q_{\theta_i}(\bm{s}_t, \bm{a}_t)$ with parameters $\theta_i (i = 1, 2)$ predict the expected reward for the pair of current state sequence $\bm{s}_t$ and action $\bm{a}_t$. The following sections will introduce the network architecture and explain how to design the loss for training $\pi_{\phi}$ and $Q_{\theta_i}$ to find the optimal parameters $\phi$ and $\theta_i$.

\subsection{Actor}
The actor consists of a GRU and two dense layers with rectified linear units (ReLU) between each layer and a final tanh unit following the output layer (see \autoref{fig:1_overview}). The actor takes the state sequence $\bm{s}_t \in \mathbb{R}^{T\times22}$ of a fixed length $T$ as the input and generates the optimal action $\bm{a}_t \in \mathbb{R}^{6}$ based on its parameters $\phi$ to update the torques or angles of six joint motors on the robot directly. To maximize the expected reward $J(\phi)$, the parameters of $\pi_{\phi}$ can be updated using the inverse gradient of the  $J(\phi)$ based on the deterministic policy gradient(DPG)  theorem\cite{silver_deterministic_2014}: 
\begin{equation}
\label{eq:actor_dpg}
\begin{split}
    \nabla_{\phi}J(\phi) &= \mathbb{E}_{
    \bm{s} \thicksim p_{\pi}}[-\nabla_a Q_{\theta}(s, a)|_{a=\pi(s)}\nabla_{\phi}\pi_{\phi}(s)], \\
    \phi &\leftarrow \phi - \frac{\alpha}{Z} \nabla_{\phi}J(\phi),
\end{split}
\end{equation}
where the $\alpha$ and $Z$ are the learning rate and the parameter to normalize the gradient, respectively.

\subsection{Adversarial critics}
\label{subsec: adversarial_critics}
The critics in this paper are composed of a GRU and three dense layers with ReLU units (see \autoref{fig:1_overview}). Each critic takes the state sequence $\bm{s}_t \in \mathbb{R}^{T\times22}$ and the action $\bm{a}_t \in \mathbb{R}^{6}$ as inputs and estimates the expected reward $J(\phi)$. The parameters of each critic are updated through the temporal difference (TD) learning based on the Bellman equation, which is a fundamental relationship between the Q-value of the current state-action pair and that of the subsequent state-action pair:
\begin{equation}
\label{eq:bellman}
\begin{split}
    Q_{\theta}(\bm{s}_t, \bm{a}_t) &= r_{t} + \gamma \mathbb{E}_{\bm{s}_{t+1}, \bm{a}_{t+1} \thicksim \pi_\phi(\bm{s}_{t+1})}[Q_{\theta}(\bm{s}_{t+1}, \bm{a}_{t+1})].\\
\end{split}
\end{equation}

To calculate the $\mathbb{E}_{\bm{s}_{t+1}, \bm{a}_{t+1}}[Q_\theta(\bm{s}_{t+1}, \bm{a}_{t+1})]$ in the \eqref{eq:bellman}, previous researchers designed a target network $Q_{\theta'}(\bm{s}, \bm{a})$ to preserve a fixed objective $y$ over multiple time steps \cite{fujimoto_addressing_2018}:
\begin{equation}
\label{eq:deep_Q_learning}
\begin{split}
    &y = r_{t} + \gamma Q_{\theta'}(\bm{s}_{t+1}, \bm{a}_{t+1}),\\
    &\bm{a}_{t+1} \thicksim \pi_{\phi'}(\bm{s}_{t+1}),
\end{split}
\end{equation}
where the new action $\bm{a}_{t+1}$ is generated from a target actor network $\pi_{\phi'}$. 


To train the critic network to estimate the target objective $y$ accurately, the loss of the critic network is usually a mean squared error between the current Q-value $Q_\theta(\bm{s}_t, \bm{a}_t)$ and the target objective $y$:
\begin{equation}
\label{eq:loss}
\begin{split}
    \mathcal{L}(\theta) = \mathbb{E}_{\pi_{\phi'}}[(Q_\theta(\bm{s}_t, \bm{a}_t)- y)^2].
\end{split}
\end{equation}

Because the above Q-value is updated with a greedy target $y = r_{t} + \gamma \max_{\bm{a}_{t+1}} Q_{\theta'}(\bm{s}_{t+1}, \bm{a}_{t+1})$ and the actor network is updated using the DPG theorem, it is possible to overestimate the Q-value \cite{fujimoto_addressing_2018}. The overestimation error can be accumulated since \eqref{eq:deep_Q_learning} may create an error feedback loop: the suboptimal critic highly rates the suboptimal action and influences the next update of the actor.

To mitigate the overestimation error, TD3 creates two independent critics $Q_{\theta_1}$ and $Q_{\theta_2}$ and two corresponding target critic networks $Q_{\theta'_1}$ and $Q_{\theta'_2}$. Then the objective is updated using the minimum value of two target Q-values:
\begin{equation}
\begin{split}
    &y = r_{t} + \gamma \min_{i=1, 2} Q_{\theta'_i}(\bm{s}_{t+1}, \bm{a}_{t+1})\\
    &\bm{a}_{t+1} \thicksim \pi_{\phi'}(\bm{s}_{t+1})
\end{split}
\end{equation}

Because two critics track the same objective $y$, they may not be independent after training many times and the effect of selecting the minimum Q-value will be weakened. To address this issue, this paper designs an adversarial loss $L_a(\theta_i)$:
\begin{equation}
\label{eq:loss_a}
\begin{split}
    &\mathcal{L}_a(\theta_i) = -\mathbb{E}_{\pi_{\phi'}}[(Q_{\theta_1}(\bm{s}, \bm{a})- Q_{\theta_2}(\bm{s}, \bm{a}))^2]\\
    &\theta_i  \leftarrow \arg \min_{\theta_i} (\mathcal{L}(\theta_i) + \beta \mathcal{L}_a(\theta_i)), i =1, 2
\end{split}
\end{equation}
where $\beta \in [0, 0.5)$ is a temperature parameter to control the importance of the adversarial loss and is set at 0.1.

The output layer for each critic $Q_{\theta_i}(i = 1, 2)$ is updated based on the gradient descent. To simplify the equation, here we ignore the input $\bm{s}$ and $\bm{a}$.
\begin{equation}
\label{eq:gradient}
\begin{split}
    Q_{\theta_i} &\leftarrow Q_{\theta_i} - \alpha \nabla_{Q_{\theta_i}} [(y - Q_{\theta_i})^2 - \beta (Q_{\theta_1} - Q_{\theta_2})^2]\\
    Q_{\theta_i} &\leftarrow Q_{\theta_i} + 2\alpha(y - Q_{\theta_i}) + 2\alpha \beta (Q_{\theta_i} - Q_{\theta_{3-i}})
\end{split}
\end{equation}
where $\alpha$ is the learning rate.

As shown in \eqref{eq:gradient}, the gradient $\beta(Q_{\theta_i} - Q_{\theta_{3-i}})$ caused by the adversarial loss increases the step size of the $Q_{\theta_i}$ if $(y - Q_{\theta_i})$ and $(Q_{\theta_i} - Q_{\theta_{3-i}})$ are of the same sign, otherwise it decreases the step size of the $Q_{\theta_i}$. When $y - Q_{\theta_1}$ and $y - Q_{\theta_2}$ are of the same sign, the adversarial loss increases the step size of one Q-value while decreases the step size of another to promote these two Q-values to distribute on different sides of the target objective $y$. After $y - Q_{\theta_1}$ and $y - Q_{\theta_2}$ are of opposite signs, the adversarial loss decreases the step size of two Q-values. Because the temperature parameter $\beta$ is lower than 0.5, the gradient $y - Q_{\theta_i}$ is dominant and lead the Q-value to converge to the targe $y$. Therefore, the adversarial loss does not prevent Q-values from converging to the target and it promotes two Q-values to distribute on different sides of the target objective $y$ during the training process. Then the minimum target Q-value will be the lower boundary of the target Q-value. This strategy may underestimate Q value, but the underestimation is preferable to overestimation because it is less possible to explicitly propagate the underestimated actions through the policy update \cite{fujimoto_addressing_2018}. Besides, the mean of Q values will be closer to the target Q-value than the single Q value and improve the performance of the learned policy.


The ATD3 uses the same hyperparameters as TD3 \cite{fujimoto_addressing_2018}. Adam optimizer with a learning rate of $10^{-3}$ is used to train the network. All transitions are saved in a buffer, \textcolor{black}{and 100 transitions are sampled from the buffer to train the network after finishing an episode. The number of training times equals to that of the time steps in the current episode.} To remove the dependency of the initial parameters, the robot adopts random actions in the first 10,000 time steps. Besides, a Gaussian noise of $\mathcal{N}(0, \sigma)$ is added on the output action to explore different situations. The ATD3 algorithm is shown in Algorithm \ref{alg:ATD3}, which is an improved version of TD3 \cite{fujimoto_addressing_2018}.

\begin{algorithm}[h!]
\caption{ATD3}
\label{alg:ATD3}
\begin{algorithmic}
   \STATE Initialize critics $Q_{\theta_1}$, $Q_{\theta_2}$ and the actor $\pi_\phi$ using random parameters $\theta_1$, $\theta_2$, $\phi$.
   \STATE Initialize target networks $\theta'_1 \leftarrow \theta_1$, $\theta'_2 \leftarrow \theta_2$, $\phi' \leftarrow \phi$.
   \STATE Initialize replay buffer $\mathcal{B}$.
   
   \FOR{$t=1$ {\bfseries to} $T$}
   \STATE Generate action with exploration noise $\bm{a}_t \thicksim \pi_\phi(\bm{s}_t) + \bm{\varepsilon}_t$, 
   \STATE $\bm{\varepsilon}_t \thicksim \mathcal{N}(0, \sigma)$, and observe reward $r_t$ and new state $\bm{s}_{t+1}$.
   \STATE Store state-action tuple $(\bm{s}_{t}, \bm{a}_{t}, r_t, \bm{s}_{t+1})$ in $\mathcal{B}$. 
   
   \IF{an episode is finished}
   \FOR{$i=1$ {\bfseries to} time steps in the episode}
   \STATE Sample mini-batch of $N$ transitions $(\bm{s}_{t}, \bm{a}_{t}, r_t, \bm{s}_{t+1})$ from $\mathcal{B}$.
   \STATE $\tilde {\bm{a}}_{t+1} \leftarrow \pi_{\phi'}(\bm{s}_{t+1}) + \bm{\varepsilon}_t, ~\bm{\varepsilon}_t \thicksim \text{clip}(\mathcal{N}(0, \tilde \sigma), -c, c)$.
   \STATE $y \leftarrow r_t + \gamma \min_{i=1,2} Q_{\theta'_i}(\bm{s}_{t+1}, \tilde {\bm{a}}_{t+1})$.
   \STATE Update critics $\theta_i \leftarrow \arg \min_{\theta_i} N^{-1} \sum[(y - Q_{\theta_i}(\bm{s}_t,\bm{a}_t))^2 - \beta (Q_{\theta_1}(\bm{s}_t,\bm{a}_t) - Q_{\theta_2}(\bm{s}_t,\bm{a}_t))^2]$.
   
   \IF{$t$ mod $d$}
   \STATE Update $\phi$ by the deterministic policy gradient $\nabla_{\phi} J(\phi)$:
   \STATE $N^{-1} \sum [-\nabla_{\bm{a}_t} \bar{Q}_{\theta}(\bm{s}_t, \bm{a}_t) |_{\bm{a}_t=\pi_{\phi}(\bm{s}_t)} \nabla_{\phi} \pi_\phi(\bm{a}_t)]$.
   \STATE $\bar{Q}_{\theta}(\bm{s}_t, \bm{a}_t) = \big({Q_{\theta_1}(\bm{s}_t, \bm{a}_t) + Q_{\theta_2}(\bm{s}_t, \bm{a}_t)}\big)/{2}$
   \STATE Update target networks:
   \STATE $\theta'_i \leftarrow \tau \theta_i + (1 - \tau) \theta'_i$,
   \STATE $\phi' \leftarrow \tau \phi + (1 - \tau) \phi'$.
   \ENDIF
   \ENDFOR
   \ENDIF
   \ENDFOR
\end{algorithmic}
\end{algorithm}

\section{Results}
\label{sec:Results}
\subsection{Ablation study of different gait rewards}
Although the sub gait rewards provided in \autoref{tab:gait_reward} seem to be plausible, some rewards may have negative effects on training the biped robot to walk. To find the optimal gait rewards, the ablation studies of the gait rewards were first implemented with the Roboschool Walker2d robot through the OpenAI Gym \cite{brockman_openai_2016}. In each trial, the Gym simulator and the actor-critic networks were initiated randomly, then the robot was trained for 0.3 million time steps and evaluated every 5000 time steps. The exploration noise was set as
$\mathcal{N}(0, 0.1)$ in the training stage. Each evaluation reported the average episode reward with no exploration noise in 10 episodes. The gait rewards designed in this paper were only added in the training process, and the evaluated reward was the sum of the default reward $\sum{r^d}$ in an episode, which allows others to repeat the experiment conveniently. The maximum number of time steps in each episode was 1000, and the simulation would stop earlier if the robot fell. 

\begin{figure}[h!]
    \includegraphics[width=\columnwidth]{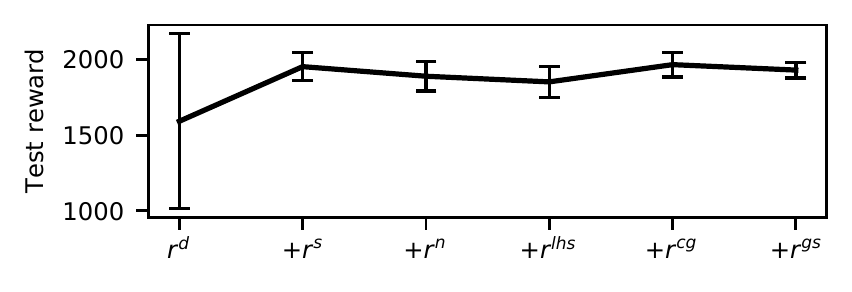}
    \caption{Mean and standard deviation values of the highest test reward using different training rewards. The reward on an $x$ coordinate and its left rewards are summed to train the robot, and the corresponding test reward $\sum{r^d}$ is calculated. The reward definitions are shown in (\ref{eq:default_reward}) and \autoref{tab:gait_reward}.}
    \label{fig:ablation_reward}
\end{figure}

In the ablation experiments, the designed rewards were summed incrementally and the TD3 was utilized to train the robot. For each combination of the rewards, five random trials were repeated. The mean and standard deviation values of the evaluation results for different rewards are shown in \autoref{fig:ablation_reward}. The results indicate that the optimal choice is the sum of the default reward $r^d$, the reward of the double support period $r^s$, the gait number $r^n$, the left heel strike $r^{lhs}$, and the crossover gait $r_{cg}$, which can be denoted by $r^d + \hat{r}^g$. The $r^d + \hat{r}^g$ may not be the best because the sequence of rewards affects the combination of the rewards in the ablation study, but it is much better than the default reward. The reward of gait symmetry $r^{gs}$ slightly negatively affect the test reward, possibly because the robot sometimes needs to sacrifice the gait symmetry and adjust the gait to rebalance.

\subsection{Test rewards using different methods}
The effects of different methods were then analyzed in the Roboschool and Webots simulators. The exploration noise and the maximum training time steps were changed to $\mathcal{N}(0, 0.2)$ and 1 million in the Webots simulator. Other experimental setups were the same in two simulators. 

\begin{figure}[h!]
    \centering
    \includegraphics[width=0.99\columnwidth]{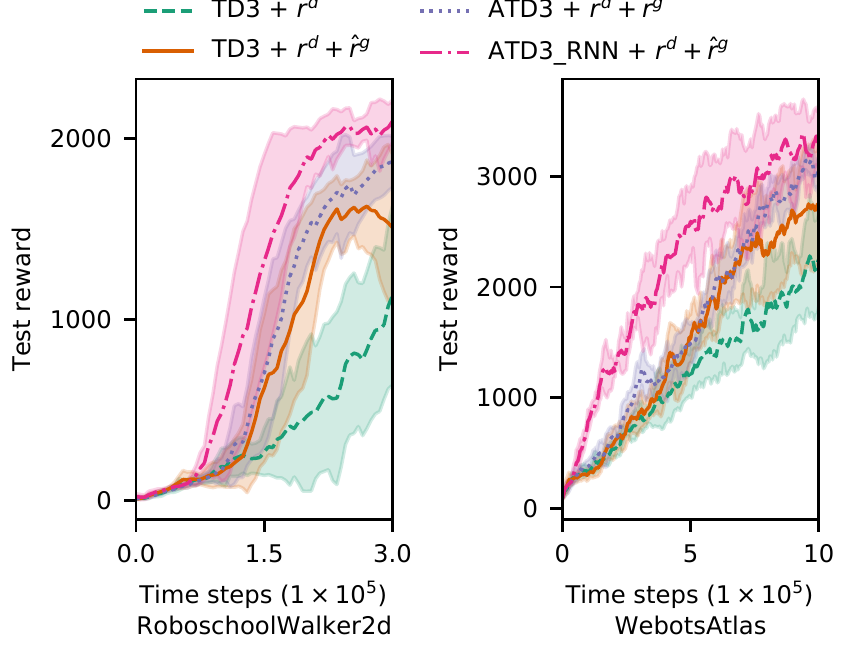}
    \caption{Learning curves for the Roboschool Walker2d and Webots Atlas tasks. The shaded region represents a standard deviation of the mean evaluation in five repeating training experiments. Curves are smoothed uniformly using an exponential moving average filter whose weight is 0.8.}
    \label{fig:test_reward} 
\end{figure}

The evaluations were compared among four different methods: using TD3 and the default reward (TD3 $+~r^d$), adding the optimal gait reward and using TD3 (TD3 $+~r^d + \hat{r}^g$), ATD3 (ATD3 $+~r^d + \hat{r}^g$), or ATD3 with RNN (ATD3\_RNN $+~r^d + \hat{r}^g$). The contrast results are illustrated in \autoref{tab:rewards_methods} and \autoref{fig:test_reward}. 

In the Roboschool Walker2d task, the highest test reward is \textcolor{black}{1592.06 $\pm$ 578.32} without the gait reward $\hat{r}^g$. The robot sometimes remains standing to obtain the alive reward. After adding the gait reward, the mean of test reward increases 23.50\% and the variance decreases a lot (see \autoref{tab:rewards_methods}). Hence, the gait reward promotes the robot to walk better. The ATD3 further improves the results. As shown in \autoref{fig:test_reward}, the test reward of ATD3 increases more stably than that of TD3, and thus the adversarial loss and the mean Q-value make ATD3 more robust. Besides, the robot obtains a higher test reward using ATD3, which is 1.43\% higher than using TD3. The test reward achieves 14.32\% higher after adding the RNN because the historical states are also utilized to estimate the Q-value and select the optimal action. The $r^d + \hat{r}^g$ and ATD3\_RNN assist the robot to achieve the highest test reward, which is 43.21\% higher than using the TD3 and the default reward. Besides, The $r^d + \hat{r}^g$ and ATD3\_RNN help to decrease the variance of the test reward, which increases the stability of the result. With lower variance, RL methods become more reproducible. However, there was a probability that the reward did not converge after using the RNN, and the probability increased with the sequence length of the RNN, possibly because the longer sequence length increased the state-action space. The sequence length of the RNN was set at 2 in this paper, which decreased the failure probability to about 5\% estimated in 20 repeating trials. 

The evaluation results are similar in the Webots Atlas task. As shown in \autoref{tab:rewards_methods}, the mean of the highest test reward using TD3 $+~r^d + \hat{r}^g$, ATD3 $+~r^d + \hat{r}^g$, ATD3\_RNN $+~r^d + \hat{r}^g$ increase \textcolor{black}{9.63\%, 17.54\%, and 23.53\%} compared to that of using TD3 $+~r^d$. As shown in \autoref{fig:RoboschoolWalker2d_walking_results}, the variance of the test reward using the ATD3 is also lower than that using the TD3, which validates that the adversarial critics and averages Q values can increase the stability of training the robot. Besides, ATD\_RNN accelerates the learning process of the Atlas robot, which is similar to the results of the Walker2d.



\begin{table}[htbp]
\centering
\caption {\label{tab:rewards_methods} Mean and standard deviation values of the highest test reward using different methods.}
\renewcommand{\arraystretch}{1} 
\begin{center}
\resizebox{0.99\columnwidth}{!}{
\begin{tabular}{l c c c c}
\toprule
Methods & Mean (\%) & SD (\%) & Mean (\%) & SD (\%)\\
\midrule
 & \multicolumn{2}{c}{Walker2d} & \multicolumn{2}{c}{Atlas}\\
\midrule
TD3 $+~r^d$ & 1592.06 & 578.32 & 3117.75 & 337.90\\
TD3 $+~r^d + \hat{r}^g$ & 1966.21 & 81.62 & 3417.88 & 447.08 \\
ATD3 $+~r^d + \hat{r}^g$ & 1994.40 & 147.55 & 3664.69 & \textbf{184.10} \\
ATD3\_RNN $+~r^d + \hat{r}^g$ & \textbf{2280.06} & \textbf{71.46} & \textbf{3851.21} & 186.52
\end{tabular}
}
\end{center}
\end{table}

\subsection{Estimated Q-value}
\begin{figure}[h!]
    \centering
    \vspace*{0.5em}
    \includegraphics[width=0.99\columnwidth]{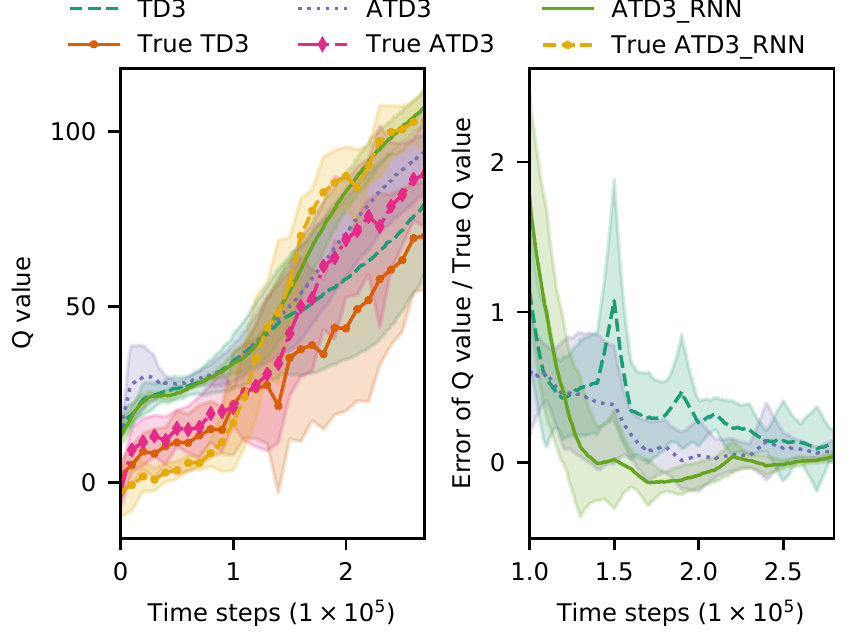}
    \caption{The estimated Q value, the true Q value, and the normalized error of the Q value. For ATD3 and ATD3\_RNN, the estimated Q value is estimated by sampling 10,000 state-action pairs from the replay buffer, computing the mean of Q values of two critics for each pair, and averaging them. The TD3 only calculates the average Q value of one critic. The true Q value is estimated by sampling 1000 state-action pairs from the replay buffer, setting them as the initial states of the environment in 1000 episodes, and calculating the average expected cumulative reward shown in (\ref{eq:expected_return}) in these 1000 episodes. The estimated Q value is calculated every 1000 time steps and the true Q value is estimated every 10,000 time steps. The normalized error of Q value is to divide the error between the estimated Q value and the true Q value by the true Q value. Five random trials are repeated to calculate their mean and standard deviation values, which are represented by the lines and shades areas, respectively.}
    \label{fig:Q_value} 
\end{figure}

TD3, ATD3, are ATD3\_RNN are updated using the deterministic policy gradients, which may induce the overestimation of the Q value and affect the final performance of the learned policy \cite{fujimoto_addressing_2018}. Section \ref{subsec: adversarial_critics} and (\ref{eq:loss_a})-(\ref{eq:gradient}) qualitatively analyzed that the adversarial critics and the average Q values can assist to estimate the true Q value more accurately. To validate this theory, the Q values estimated by the critics and the true Q values are compared among the above three methods in the Roboschool simulator. The training reward and test reward should be the same for evaluating the Q value, and thus they were set as the default reward with an offset $r^d - 0.5$ to decrease the probability of remaining standing. Besides, the trial was repeated and Q values were revaluated if the robot remained standing after training. As shown in \autoref{fig:Q_value}, The above three methods can all decrease the error between the estimated Q value and the true Q value gradually. The ATD3 can decrease the normalized error of Q value (lower mean) and more stably (lower standard deviation) than the TD3. The mean of the absolute normalized error of the Q value in the last 50,000 time steps for the TD3, ATD3, and ATD3\_RNN are \textcolor{black}{19.86\%, 9.99\%, and 3.35\%,} respectively. Hence, the adversarial critics and the average Q values improve the estimated result of the Q value, which may explain that the test rewards in two simulators increased after using the ATD3 and ATD3\_RNN.

\subsection{Gait similarity between robots and human}
\begin{figure}[h!]
    \centering
    \includegraphics[width=\columnwidth]{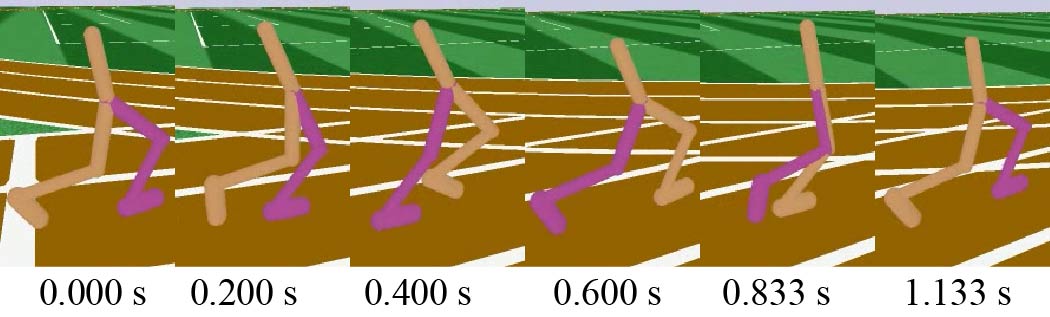}
    \caption{The gait learned by the Roboschool Walker2d.}
    \label{fig:RoboschoolWalker2d_walking_results} 
\end{figure}

\begin{figure}[h!]
    \centering
    \includegraphics[width=\columnwidth]{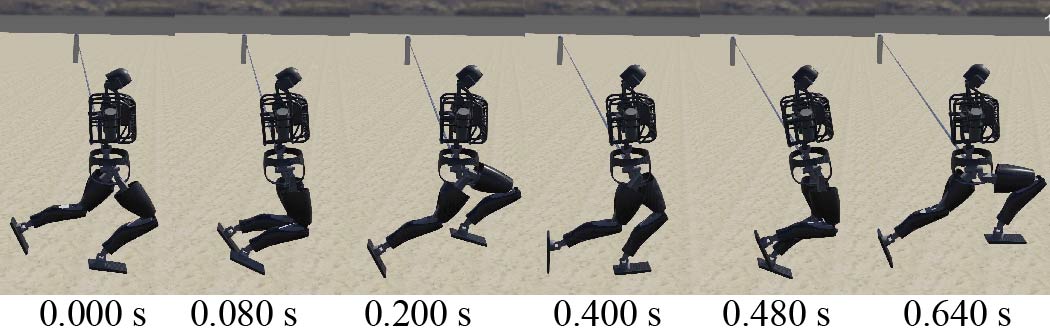}
    \caption{The gait learned by the Webots Atlas.}
    \label{fig:WebotsAtlas_walking_results} 
\end{figure}

The animation of the walking results trained by ATD3\_RNN $+~r^d + \hat{r}^g$ are visualized in \autoref{fig:RoboschoolWalker2d_walking_results} and \autoref{fig:WebotsAtlas_walking_results}. The Walker2d is able to achieve one complete gait in 1.13 s, and the gait cycle of the Atlas is about 0.64 s. The gait learned by the Walker2d and Atlas are similar to the walking and running gait of human, respectively.

\begin{figure}[h!]
    \centering
    \vspace*{0.5em}
    \includegraphics[width=0.99\columnwidth]{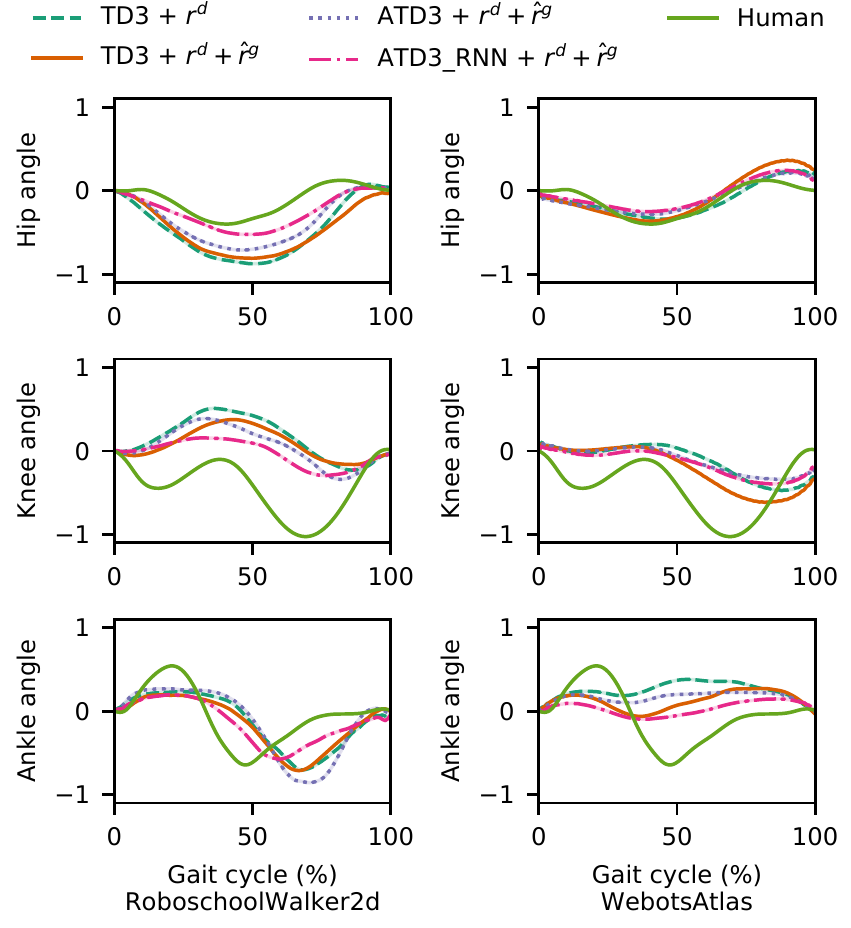}
    \caption{Joint angles in a gait produced by different methods. Joint angles of hip, knee, and ankle are normalized from [-45$^\circ$, 115$^\circ$], [150$^\circ$, 0$^\circ$], and [-45$^\circ$, 45$^\circ$] to [-1, 1] and subtracted by the initial angle. }
    \label{fig:joint_angle} 
\end{figure}

A purpose of setting a gait reward is to teach the robot to learn bipedalism from the walking principles of humans. To illustrate the learned results of the biped robot, the joint angle of the hip, knee, and ankle are shown in \autoref{fig:joint_angle}. The saved policies in 5 random training trials were utilized to control the robot. Then the trajectory of the robot was segmented into different gaits based on the foot strike event. The gait starts at the foot strike event and ends at the next ipsilateral foot strike event. The joint angles in a gait may have different lengths and were re-sampled to the length of 100. Then the mean values of the joint angles in a gait were calculated and plotted in \autoref{fig:joint_angle}. We also calculated the cosine similarity between joint angles of human and those of robot, and the similarity results for TD3 $+~r^d$, TD3 $+~r^d + \hat{r}^g$, ATD3 $+~r^d + \hat{r}^g$, and ATD3\_RNN $+~r^d + \hat{r}^g$ are 0.34, 0.38, 0.45, and 0.71 for the Roboschool Walker2d robot, and 0.39, 0.54, 0.46, and 0.66 for the Webots Atlas robot, respectively. The Walker2d and Atlas using ATD3\_RNN $+~r^d + \hat{r}^g$ achieves 108.82\% and 69.23\% higher similarity than using TD3 $+~r^d$. Therefore, the gait reward and proposed ATD3\_RNN increases the kinematic similarity between human and robot, even if there is no human joint angle nor handcrafted reference trajectory.


\section{Conclusion}
\label{sec:Conclusion} 
\textcolor{black}{Maximizing the cumulative reward needed to avoid the local minima and reduce the error of estimating the cumulative reward.} However, most end-to-end RL methods did not consider the prior knowledge of the biped walking, which may lead the robot to fall into the local minima and learn strange motions. Additionally, the value-based RL may overestimate the cumulative rewards. \textcolor{black}{To increase the cumulative reward, this paper proposed a gait reward and an ATD3\_RNN algorithm. The gait reward was set based on walking principles and assisted the robot to escape from the local minima (e.g., standing). The ATD3\_RNN further increased the cumulative reward by decreasing the error of estimating the Q value via designing two adversarial critics, averaging Q values, and adding an RNN.} The proposed gait reward and algorithm were evaluated on two robots: Roboschool Walker2d and Webots Atlas. \textcolor{black}{The experimental results showed that the ATD3\_RNN decreased the error of estimating the expected reward from 19.86\% to 3.35\%.} The proposed gait reward ($\hat{r}^g$) and ATD3\_RNN led to 43.21\% and 23.53\% higher reward than only using the TD3 and the default reward on the two robots. The cosine similarity between human gait and the motion generated by the Walker2d and Atlas with ATD3\_RNN $+~r^d + \hat{r}^g$ were 108.82\% and 69.23\% higher than those with TD3 $+~r^d$. \textcolor{black}{Hence, the proposed reward and ATD3\_RNN reduced the situations of the local minima and decreased the error of estimating the cumulative reward, and thus increased the cumulative reward and taught the robot to walk better.}
\bibliographystyle{IEEEtran}
\bibliography{root}
\end{document}